\documentclass[conference]{IEEEtran}
\IEEEoverridecommandlockouts

\usepackage{hyperref}
\usepackage{cite}
\usepackage{amsmath,amssymb,amsfonts}
\usepackage{algorithmic}
\usepackage{graphicx}
\usepackage{textcomp}
\usepackage{xcolor}
\usepackage{colortbl}

\usepackage{fixmath}
\usepackage{mathtools}

\usepackage[acronym]{glossaries}
\usepackage{csquotes}

\usepackage{enumitem}

\usepackage{url}
\usepackage{multirow}

\usepackage{float}                  
\usepackage[figuresright]{rotating} 
\usepackage{subcaption}             
\usepackage{eso-pic}                
\usepackage{overpic}                
\usepackage{tikz,pgfplots,pgfkeys}  
\usepackage{psfrag}                 
\usepackage{tabulary}
\usepackage{xparse}                 
\usepackage{todonotes, wasysym} 	  

\usepackage[nodisplayskipstretch]{setspace}

\newcommand{\norm}[1]{\left\| #1 \right\|_2}

\renewcommand{\vec}[1]{\ensuremath{\boldsymbol{\lowercase{#1}}}}

\newcommand{\real}{\mathbb{R}}

\newcommand{\loss}{\ensuremath{\mathcal{L}}}

\renewcommand{\^}[1]{\ensuremath{^{(\text{#1})}}}

\usepackage{pdfpages}
\usepackage[most]{tcolorbox}

\usepackage{xparse}

\def\exampletext{\small Example} 
\NewDocumentEnvironment{testexample}{ O{} }
{
\colorlet{colexam}{black} 
\newtcolorbox[use counter=testexample]{testexamplebox}{%
    empty,
    size=minimal,
    title={\small \exampletext~\thetcbcounter: #1},
    attach boxed title to top left,
       minipage boxed title,
    boxed title style={empty,size=minimal,toprule=0pt,top=2pt,left=-2mm,overlay={}}, 
    coltitle=colexam,fonttitle=\bfseries,
    before=\par\medskip\noindent,parbox=false,boxsep=0pt,left=-2.3mm,right=0mm,top=2pt,breakable,pad at break=0mm,
       before upper=\csname @totalleftmargin\endcsname0pt, 
    overlay unbroken={\draw[colexam,line width=.75pt] ([xshift=-8pt, yshift=-1pt]title.north west) -- ([xshift=-8pt, yshift=-2pt]frame.south west); },
    }
\begin{testexamplebox}}
{\end{testexamplebox}\endlist}

\NewDocumentEnvironment{testexample2}{ O{} }
{
\colorlet{colexam}{black} 
\newtcolorbox[use counter=testexample2]{testexamplebox2}{%
    empty,
    size=minimal,
    title={\small Review~\thetcbcounter : #1},
    attach boxed title to top left,
       minipage boxed title,
    boxed title style={empty,size=minimal,toprule=0pt,top=2pt,left=1.3mm,overlay={}}, 
    coltitle=colexam,fonttitle=\bfseries,
    before=\par\medskip\noindent,parbox=false,boxsep=0pt,left=1.3mm,right=0mm,top=2pt,
       before upper=\csname @totalleftmargin\endcsname0pt, 
    overlay unbroken={\draw[colexam,line width=.75pt] ([xshift=1.6pt, yshift=-1.2pt]title.north west) -- ([xshift=1.6pt, yshift=-1.2pt]frame.south west); },
    }
\begin{testexamplebox2}}
{\end{testexamplebox2}\endlist}

\newcommand{\authorA}{\includegraphics[width=1.8ex]{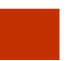}}
\newcommand{\authorB}{\includegraphics[width=2.0ex]{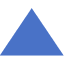}}

\newcommand{\subparagraph}{}
\usepackage{titlesec}
\titlespacing*{\section}{0pt}{1.9ex}{.1ex}
\titlespacing*{\subsection}{0pt}{1.2ex}{.1ex}
\titlespacing*{\subsubsection}{0pt}{.1ex}{.01ex}

\IEEEoverridecommandlockouts
\IEEEpubid{\makebox[\columnwidth]{\copyright2019
IEEE \hfill} \hspace{\columnsep}\makebox[\columnwidth]{ }} 

\begin{document}

\setlength{\belowdisplayskip}{4.pt} \setlength{\belowdisplayshortskip}{4.5pt}
\setlength{\abovedisplayskip}{4.pt} \setlength{\abovedisplayshortskip}{4.5pt}

\title{Explainable Authorship Verification in Social Media via Attention-based Similarity Learning
\thanks{This work was in significant parts performed on a high performance computing cluster at
 Bucknell University through the support of the National Science Foundation, Grant Number 1659397.}
}

\author{
\IEEEauthorblockA{Benedikt Boenninghoff$~^1$, Steffen Hessler$^2$, Dorothea Kolossa$^1$, Robert M. Nickel$^3$}
\IEEEauthorblockA{
        $^1$\textit{Cognitive Signal Processing Group, Ruhr University Bochum, Germany}\\
        $^2$\textit{Department of German Philology , Ruhr University Bochum, Germany}\\
        $^3$\textit{Department of Electrical and Computer Engineering, Bucknell University, Lewisburg, PA, USA} \\
        Emails: benedikt.boenninghoff@rub.de,
        steffen.hessler@rub.de,
        dorothea.kolossa@rub.de, rmn009@bucknell.edu
        \vspace*{-.3cm}}
}

\maketitle

\newacronym{CNNs}{CNNs}{convolutional neural networks}
\newacronym{RNN}{RNN}{recurrent neural network}
\newacronym{LSTM}{LSTM}{long-short term memory}
\newacronym{MLP}{MLP}{multilayer perceptron}
\newacronym{SVM}{SVM}{support vector machine}
\newacronym{HRSN}{HRSN}{hierarchical recurrent Siamese network}

\begin{abstract}
Authorship verification is the task of analyzing the linguistic patterns of two or more texts to determine whether they were written by the same author or not. The analysis is traditionally performed by experts who consider \textit{linguistic} features, which include spelling mistakes, grammatical inconsistencies, and stylistics for example.
Machine learning algorithms, on the other hand, can be trained to accomplish the same, but have traditionally relied on so-called \textit{stylometric} features.
The disadvantage of such features is that their reliability is greatly diminished for short and topically varied social media texts. In this interdisciplinary work, we propose a substantial extension of a recently published hierarchical Siamese neural network approach, with which it is feasible to learn \textit{neural} features and to visualize the decision-making process. 
For this purpose, a new large-scale corpus of short Amazon reviews for text comparison research is compiled and we show that the Siamese network topologies  outperform state-of-the-art approaches that were built up on stylometric features. Our linguistic analysis of the internal attention weights of the network shows that the proposed method is indeed able to latch on to some traditional linguistic categories.

\end{abstract}

\begin{IEEEkeywords}
Authorship verification, similarity learning, forensic text comparison, Siamese network, deep metric learning
\end{IEEEkeywords}

\vspace*{-0.5cm}
\section{Introduction}
\label{sec:intro}

\begin{figure}[t!]
\vspace*{-.3cm}
\centering
\small
\begin{testexample2}[$\boldsymbol{y}_1$]
        \scalebox{0.86}{
        \input{differentialanalysis/doc_L0.tex}
        }
\end{testexample2}
\vspace*{-.2cm}
\begin{testexample2}[$\boldsymbol{y}_2$]
        \scalebox{0.86}{
            \input{differentialanalysis/doc_R0.tex}
            }
\end{testexample2}
\vspace*{-.2cm}
\caption{Attention-heatmap of our proposed authorship verification framework. Blue hues encode the sentence-based attention weights and red hues denote the relative word importance. All tokens are delimited by whitespaces.}
\label{fig:example}
\vspace*{-.6cm}
\end{figure}

In authorship verification, also known as (forensic) text comparison, two or more text documents are compared with respect to their style to ascertain if the documents were written by the same author or not. Authorship verification is typically performed by experts who rely on traditional linguistic categories in their analysis. These categories include peculiarities of spelling/grammar, stylistic mannerisms, dialects, sociolects, and registers of language that hint at the authorship of a disputed document. Even though there is no such thing as a ``linguistic fingerprint,'' the linguistic features people exhibit in their writing are specific enough to be admitted as evidence in court. 
The Federal Criminal Police (Bundeskriminalamt) in Germany, for example, uses text comparison techniques to identify potential suspects and/or to substantiate charges against a suspect in a criminal case~\cite{BKA}.
A comprehensive overview of commonly employed categories in forensic linguistic can be found in~\cite{mcmenamin2002forensic}.
Important practical implications of forensic linguistic, such as the relationship between language, criminal justice, and the law are explored in~\cite{articleLawrence}.
Research results on so-called idiolectal linguistic features that have become highly influential to authorship analysis and verification in general are reported in~\cite{10.1093/applin/25.4.431}. 
A particularly insightful description of the particular language used in fraud cases was provided in~\cite{shuy2016language}.
In addition, authorship verification is of great interest not only for the collection of evidence in criminal investigations, but also for the detection of deceptive intent and fake news in e-commerce and social media.

Especially social media platforms have become an ubiquitous way of communication. Unfortunately, these platforms are also notorious for the proliferation of information from unverified sources. As a result, users can fall victim to criminal predators through misinformation, fraud, and identity theft. The verification of the authorship of a piece of information can help to reduce the threat. The data volume shared on social media platforms on a daily basis, however, makes it utterly infeasible to rely on trained linguists for the analysis. 
Engineers and computer scientists have thus begun to automate parts of this process~\cite{7555393}.

From a technical point of view, we can roughly distinguish between automatic authorship attribution and automatic authorship verification.
The term \textit{authorship attribution} describes a traditional classification task. Given a finite set of candidate authors, the objective is to determine who, from a set of enrolled authors, has written a document of unknown authorship~\cite{Juola:2006:AA:1373450.1373451}. 
In contrast, the objective of \textit{authorship verification} or \textit{text comparison} is to determine whether two separate documents were written by the same author~\cite{Koppel04, doi:10.1002/asi.22954}.

Methods for authorship analysis have traditionally been based on the extraction of stylometric features
~\cite{luyckx-daelemans-2008-authorship, localhist, Afroz12, 6638728, Sapkota15, Halvani16, 
doi:10.1002/asi.24183}.
Stylometric features can generally be categorized into several distinct groups, e.g. lexical features, character features, syntactic features, semantic features, and application-specific features or compression-based features~\cite{838202, Stamatatos09, 8116753}.

In contrast to stylometric-feature-based systems, there have also been a number of relatively recent papers that 
integrate the feature extraction task into a deep learning framework for authorship attribution
~\cite{W17-4907, E17-2106} and 
verification~\cite{articlebagnall, DBLP:conf/simbig/Litvak18}.

\begin{figure}[!t]
    \centering
    \begin{psfrags}
    \newcommand{\m}{.9}
    \newcommand{\n}{.75}
    \psfrag{A}[c][c][\m]{Genre}
    \psfrag{B}[c][c][\m]{Writing style}
    \psfrag{C}[c][c][\m]{Topic}
    \psfrag{a1}[c][c][\n]{linguistic}
    \psfrag{a2}[c][c][\n]{registers}
    \psfrag{b1}[c][c][\n]{Obfuscation}
     \psfrag{b2}[c][c][\n]{Imitation}
    \psfrag{b3}[c][c][\n]{Dialects}
    \psfrag{c1}[c][c][\n]{Sentiment}
    \psfrag{c3}[c][c][\n]{Fake news}
    \psfrag{c4}[c][c][\n]{Hate speech}
    \psfrag{c2}[c][c][\n]{Stance}
    \psfrag{p}[c][c][\n]{influenced by}
    \centerline{\includegraphics[width=5.5cm]{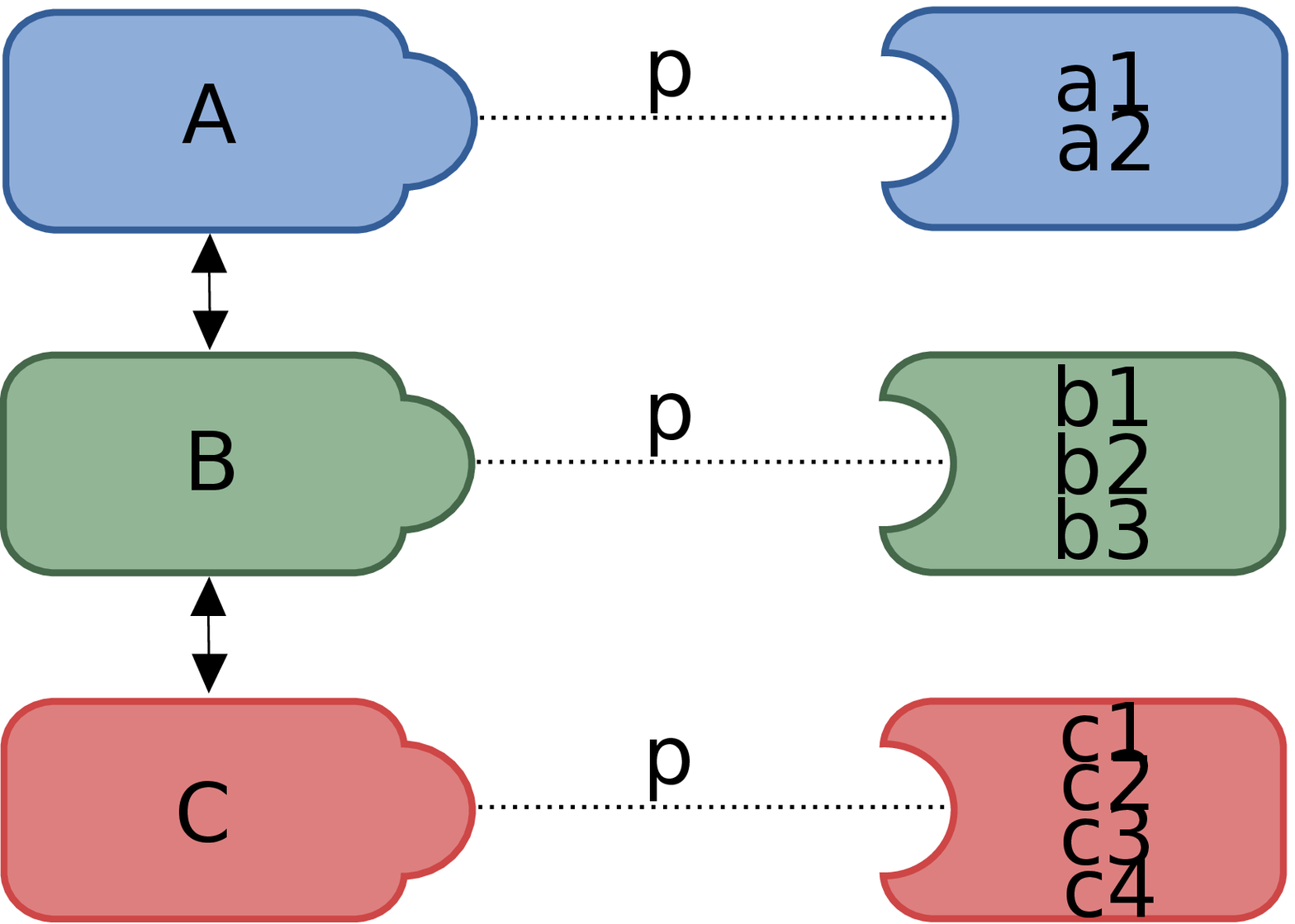}}
    \end{psfrags}
   \caption{Influences for an author's writing style.}
    \label{fig:disaffe}
    \vspace*{-.6cm}
\end{figure}

With the advent of machine learning techniques, great strides have been made in the area of authorship verification by machine. The analysis of social media texts, however, still remains challenging~\cite{8683747}. Social media texts are often short, with a high variability in genre and topical content. 
The general difficulties that arise in authorship verification for social media are best illustrated with an example from the dataset that we are considering in Section III. Fig.~\ref{fig:example} shows two product reviews from the Amazon e-commerce site. 
If we ignore the color coding for now (which will be explained later), we may quickly suspect that both reviews were written by the same author. 
Most prominently, we may perceive the particularly strong political stance implied in both texts.
In addition, we may pick up on a few repetitive patterns.
For instance, the author introduces 
a particular catch phrase and immediately follows it up with an ``explanation'' in parentheses, i.e. \textit{``I want Trump re-elected (not really !)''} in the first review and \textit{``capitalist drones (mid level managers)''} in the second review. Characteristic, also, is that the writing does not adhere to strict grammatical and spelling conventions (e.~g. \textit{``cuz''} used in the first review, third line in Fig.~\ref{fig:disaffe}, which is an alternatively spelled abbreviation:  \textit{because} $\rightarrow$ \textit{cause} $\rightarrow$ \textit{cuz})
but rather violates these rules in very idiosyncratic ways, which makes it quite difficult to rely on part-of-speech tagging for analysis~\cite{GimpelPOS}, for example. In addition, we may note that both texts exhibit strong similarities even though there is only a very limited overlap in the employed vocabulary. All of these aspects are readily noticeable for human beings, but imply significant challenges for machines. Earlier approaches have therefore had only limited success with social media data~\cite{schwartz-etal-2013-authorship}.

An additional problem that affects authorship verification algorithms is that authors tend to dramatically shift the characteristics of their writing according to the situation they are in and according to the purpose they are envisioning for the text. Three specific examples are illustrated in Fig.~\ref{fig:disaffe}:
\vspace*{-0.08in}
\begin{itemize}[leftmargin=5mm, noitemsep]
\item The \textit{writing style} can be characterized by deviations from the standard writing style of a language. Stylistic variations may appear on all branches of linguistics (e.g. syntax, morphology, phonology, lexis, semantics). In order to automatically quantify deviations from the standard language, the text collections to be examined must be sufficiently long. The writing style can be influenced, for example, by imitation or obfuscation strategies.
\item The writing style is generally also affected by the form of the text, i.e. whether we are dealing with a blackmail note, an Amazon review, a Tweet or a WhatsApp message. It has become customary, at least in parts of the literature, to refer to the type of a document generically as its {\em genre.} People change their linguistic register depending on the genre that they write in. This, in turn, leads to significant changes in computer linguistic characteristics of the resulting text. For a technical system it is thus extremely difficult to establish a common authorship between a WhatsApp message and a formal job application for example. It is therefore important to train classifiers only on one genre at a time.
\item The \textit{topic} is defined by the content of a text or the message that a person tries to communicate to the reader. The vocabulary that is used in a text tends to be strongly determined by the topic. Consequently, when the topic changes, then we observe a commensurate change in the vocabulary and thereby also a commensurate change in the characteristics of the text. It is thus desirable to develop authorship verification systems that do not put too much weight on similarity in vocabulary if we are dealing with cross-topic texts.
\end{itemize}
\vspace*{-0.08in}
In many cases only very little text is (cumulatively) available from a respective author, which poses a great challenge for the training of any deep machine learning tool. 
In addition, machine learning tools that are trained in an unrestrained fashion tend to favor topical similarity between two texts over authorship similarity. This can lead to misclassifications if two texts from the same author treat different topics, or if two documents from different authors treat the same topic. 

Motivated by this fact, we propose a novel neural network topology for the text comparison task that is applicable to a big data collection of social media texts. 
The technical core of our approach is implemented by an
\textit{\underline{A}ttention-based \underline{D}eep \underline{H}ierarchical c\underline{O}nvolutional sia\underline{M}ese b\underline{I}directional recurre\underline{N}t n\underline{E}ural-network \underline{M}odel} (\textsc{AdHominem}).

We specifically propose a substantial extension of our previous work presented in \cite{HRSN}, where a hierarchical recurrent Siamese network (\textsc{Hrsn}) was applied to encode an entire document into a single vector. 
Authorship analysis generally is an interdisciplinary field in which it is not only important to establish \textit{who} the suspected author is, but equally important to establish \textit{how} this decision was arrived at. The question of \textit{how} is crucial if we should ever hope to have neural network-based methods stand up in court. From this aspect, our previous framework (\textsc{Hrsn}) has two drawbacks: It does not use linguistically interpretable stylometric features, and it is not able to directly visualize the decision-making process. In contrast, \textsc{AdHominem} has the capacity to automatically learn meaningful neural features from a big data corpus that latch on to some traditional linguistic categories.

\begin{figure}[!t]
\centering
\begin{psfrags}
\newcommand{\n}{.8}
\psfrag{d}[c][c][\n]{distance function $d(\boldsymbol{y}_1, \boldsymbol{y}_2)$}
\psfrag{tr}[c][c][\n]{$d(\boldsymbol{y}_1, \boldsymbol{y}_2) 
                    \underset{\text{same author}}{\overset{\text{different authors}}{\gtrless}} \text{threshold $\tau$}$
                    }
\psfrag{y1}[c][c][\n]{$\boldsymbol{y}_1$}
\psfrag{y2}[c][c][\n]{$\boldsymbol{y}_2$}
\psfrag{x3}[c][c][\n]{}
\psfrag{x4}[c][c][\n]{}
\psfrag{n1}[c][c][\n]{feature extraction}
\psfrag{n2}[c][c][\n]{}
\psfrag{d1}[c][c][\n]{text 1}
\psfrag{d2}[c][c][\n]{text 2}
\psfrag{x1}[c][c][\n]{pseudo-metric}
\psfrag{x2}[c][c][\n]{learning}
\psfrag{p1}[c][c][\n]{text}
\psfrag{p2}[c][c][\n]{preprocessing}
\includegraphics[width=6cm]{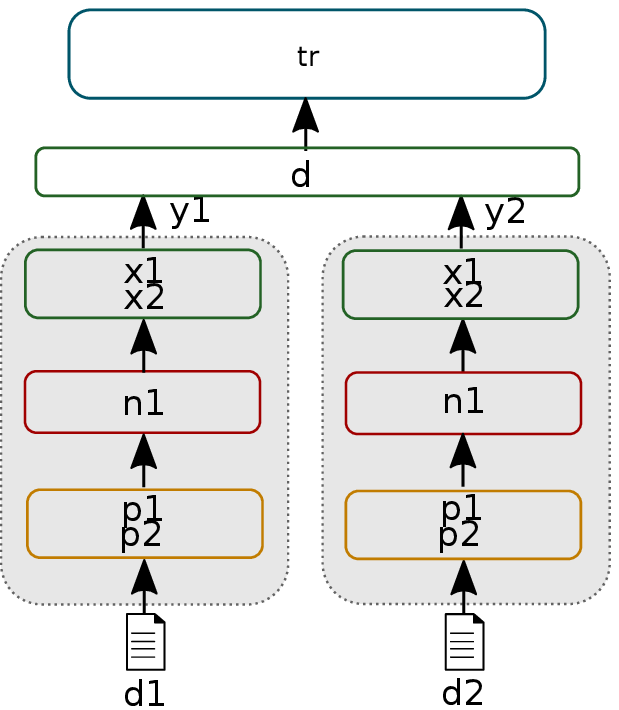}
\end{psfrags}
\caption{Overview of the proposed method \textsc{AdHominem}.}
\label{fig:overviewSiamese}
\vspace*{-.6cm}
\end{figure}

As illustrated in Fig.~\ref{fig:overviewSiamese}, our \textsc{AdHominem} topology includes three important stages: The first stage contains a text preprocessing step. The second stage includes a feature extraction in which we encode each document consisting of characters, words, and sentences into one single \textit{neural feature vector}, denoted by $\boldsymbol{y}_i$ for $ i\in {1,2}$~\cite{Yang16}. We incorporate a characters-to-word encoding layer~\cite{Ma16} to take the specific uses of prefixes and suffixes as well as spelling errors into account. Additionally, an incorporation of attention layers~\cite{Bahdanau14} allows us to visualize words and sentences that have been marked as \textit{highly significant} by the system. In the third stage, we employ a module for nonlinear metric learning to measure the similarity between two documents~\cite{Mueller16}.

We, furthermore, defined/constructed a new large dataset based on Amazon reviews for the study of authorship verification tasks. Amazon reviews do not represent forensic texts from a law enforcement/criminal point of view. We argue, nevertheless, that one can still gain valuable insights from the data into how to accomplish forensic text analysis in general. The advantages of the defined dataset are: 
\vspace*{-0.08in}
\begin{itemize}[leftmargin=5mm, noitemsep]
    \item A very large amount of reviews is publicly available and, to our best knowledge, there exists no comparable large-scale dataset containing forensic texts.
    \item Social media texts usually contain many \textit{``easy-to-visualize''} linguistic features such as spelling errors, which are crucial for forensic text comparison in general. 
    \item The provided social media corpus is annotated w.r.t authorship and topical category. Hence, by fixing the genre (see Fig.~\ref{fig:disaffe}), we are able to analyze to which degree the authorship decision is influenced by the topic. 
    \item For Amazon reviews, the number of authors trying to obfuscate their authorship is vanishingly small and obfuscation is, thus, not likely to bias our analysis of the context sensitivity. 
    \item With our proposed model, it is straightforward to incorporate a domain adaptation using smaller-sized forensic datasets, so we can easily adapt \textsc{AdHominem} to forensic text comparison.
\end{itemize}
\vspace*{-0.08in}
In summary, this paper provides the following contributions:  
\vspace*{-0.08in}
\begin{enumerate}[leftmargin=5mm, noitemsep]
 \item We propose a novel attention-based Siamese network topology with applications in authorship verification for large social media datasets.
 \item We have prepared a new large-scale corpus of short Amazon reviews for authorship verification research.
 \item In addition to a quantitative evaluation of the proposed method we present a qualitative linguistic analysis of visualized attention weights and provide evidence that the visualized attentions can provide an explanation for the decision of the network.
\end{enumerate}

\setlength{\skip\footins}{5pt}

\section{Attention-based Siamese Network Topology}
\label{sec:format}
With our \textsc{AdHominem} approach we propose a significant extension to our model introduced in~\cite{HRSN}. Its Siamese topology consists of two identical neural networks that share the exact same set of parameters. 
The network is trained to extract document-specific features to make a similarity analysis between two documents as reliable and robust as possible. 
The overall architecture is depicted in Fig.~\ref{fig:sn_att}. 
Stage one, i.e.~the preprocessing of the input texts, is described in Section~\ref{sec:impldet}. Details of stages two and three are discussed below.


\subsection{Characters-to-word Encoding}
Let $\vec{x}_{i,j,k}\^{c}$ be the real-valued $D_{\text{c}}$-dimensional character embedding vector that corresponds to the \mbox{$i$-th} character of the $j$-th word in the $k$-th sentence. We concatenate $h$ character embeddings to form a $D_{\text{c}} \cdot h$-dimensional vector
$\vec{x}\^{c}_{i:i+h-1, j, k} = \vec{x}\^{c}_{i, j, k} \oplus \vec{x}\^{c}_{i+1, j, k} \oplus \ldots \oplus \vec{x}\^{c}_{i+h-1, j, k}$ and 
apply one-dimensional convolution,
\begin{align}
    \label{eq:1DCNN}
    \boldsymbol{c}_{i,j,k} = \text{tanh}\big( \boldsymbol{W}\^{c} \vec{x}\^{c}_{i:i+h-1, j, k} + \vec{b}\^{c} \big),
\end{align}
where $\oplus$ defines the concatenation operator~\cite{Kim14}.
In Eq.~(\ref{eq:1DCNN}), the set $\boldsymbol{\theta}\^{CNN} =\{ \boldsymbol{W}\^{c} \in \real^{D_{\text{r}} \times h \cdot D_{\text{c}}}$, $\boldsymbol{b}\^{c} \in \real^{D_{\text{r}} \times 1} \}$ represents trainable parameters. Applying max-over-time pooling w.r.t.~all $D_{\text{r}}$-dimensional vectors $\boldsymbol{c}_{i,j,k}$ results in
\begin{align}
        \boldsymbol{r}_{j,k} = \max_{1 \le i \le T\^{s} - h + 1} \big\{ \boldsymbol{c}_{i, j, k}  \big\},
\end{align}
where $\boldsymbol{r}_{j,k} \in  \real^{D_{\text{r}} \times 1}$ denotes the \textit{character representation} 
of the $j$-th word in the $k$-th sentence~\cite{Ma16}. 

\subsection{Words-to-sentence Encoding}
Let $\vec{x}_{j,k}\^{w} \in \real^{D_{\text{w}} \times 1}$ be the word embedding of the $j$-th word in the $k$-th sentence. 
We now feed concatenations of character and word representations into the bidirectional LSTM network. 
The forward path at time $j \in \{1,\ldots , T\^{w}\}$ can be written as
\begin{align}
   \big\{\overrightarrow{\vec{h}}_{j,k}\^{w}, \overrightarrow{\vec{c}}_{j,k}\^{w}\big\} 
   =\text{LSTM}_{\boldsymbol{\theta}_{\text{ws}}\^{fRNN}}\bigg(\begin{bmatrix}
                                \vec{x}\^{w}_{j,k} \\ \vec{r}_{j,k} 
                             \end{bmatrix},
                              \overrightarrow{\vec{h}}_{j-1,k}\^{w}, \overrightarrow{\vec{c}}_{j-1,k}\^{w}\bigg),
\end{align}
where the hidden state and the memory state are denoted by $\overrightarrow{\vec{h}}_{j, k}\^{w}\in \real^{D_{\text{s}} \times 1}$ and 
$\overrightarrow{\vec{c}}_{j, k}\^{w}\in \real^{D_{\text{s}} \times 1}$, respectively. 
The set $\boldsymbol{\theta}_{\text{ws}}\^{fRNN}$ is comprised of all trainable 
parameters of the forward LSTM cell. 
Analogously, we denote the parameter set of the backward path with $\boldsymbol{\theta}_{\text{ws}}\^{bRNN}$. 
The joint hidden state is given by the concatenation of the forward and backward states, i.e. 
$\vec{h}_{j, k}\^{w}
                = \overrightarrow{\vec{h}}_{j, k}\^{w} \oplus  \overleftarrow{\vec{h}}_{j, k}\^{w}$. According to~\cite{Yang16} we incorporate an attention layer in the form
\begin{equation}
 \begin{split}
        \label{eq:att_w}
        \alpha_{j,k}\^{w} &= \frac{\text{exp}\big\{\vec{v}_{\text{ws}}\^{a} \vec{u}_{j,k}\^{w})\big\}}
                    {\sum_{j'=1}^{T\^{w}} \text{exp}\big\{\vec{v}_{\text{ws}}\^{a} \vec{u}_{j',k}\^{w}\big\}} \\
        \vec{x}_{k}\^{s} &= \sum_{j=1}^{T\^{w}}\alpha_{j,k}\^{w}   \vec{h}_{j, k}\^{w},        
\end{split}
\end{equation}
where  $\vec{u}_{j,k}\^{w} = \text{tanh}\big( \boldsymbol{W}_{\text{ws}}\^{a} \vec{h}_{j, k}\^{w} + \vec{b}_{\text{ws}}\^{a} \big)$ and
$\vec{x}_{k}\^{s}$ denotes the $k$-th \textit{sentence embedding}.
Trainable parameters are $\boldsymbol{\theta}_{\text{ws}}\^{ATT} = \{\boldsymbol{W}_{\text{ws}}\^{a} 
                        \in \real^{D_{\text{ws}}\^{a} \times 2 \cdot D_{\text{s}}}, 
\boldsymbol{b}_{\text{ws}}\^{a}\in \real^{D_{\text{ws}}\^{a} \times 1 } 
,\vec{v}_{\text{ws}}\^{a} \in \real^{1 \times  D_{\text{ws}}\^{a}}
\} $.


\subsection{Sentences-to-document Encoding}
On the next tier, we feed the obtained sentence embeddings into another bidirectional LSTM cell. The forward path at the $k$-th time step with $k \in \{1,\ldots , T\^{s}\}$ is given by
\begin{align}
    (\overrightarrow{\vec{h}}_{k}\^{s}, \overrightarrow{\vec{c}}_{k}\^{s}) 
            =\text{LSTM}_{\boldsymbol{\theta}_{\text{sd}}\^{fRNN}}\big(\vec{x}_{k}\^{s}, \overrightarrow{\vec{h}}_{k-1}\^{s}, \overrightarrow{\vec{c}}_{k-1}\^{s}\big),
\end{align}
where sentence-based hidden and memory states are given by 
$\overrightarrow{\vec{h}}_{k}\^{s}\in \real^{D_{\text{d}} \times 1}$ and \mbox{$\overrightarrow{\vec{c}}_{k}\^{s}\in \real^{D_{\text{d}} \times 1}$}. 
The set $\boldsymbol{\theta}_{\text{sd}}\^{fRNN}$ is comprised of all trainable 
parameters of the forward LSTM cell.
Again, we denote the parameter set of the backward path with $\boldsymbol{\theta}_{\text{sd}}\^{bRNN}$ and the combined/joint hidden state with 
$\vec{h}_{k}\^{s}
                = \overrightarrow{\vec{h}}_{k}\^{s} \oplus  \overleftarrow{\vec{h}}_{k}\^{s}$.
The attention layer at the sentence-level is then defined analogously to~(\ref{eq:att_w}),
\begin{equation}
 \begin{split}
        \label{eq:att_s}
        \alpha_{k}\^{s} &= \frac{\text{exp}\big\{\vec{v}_{\text{sd}}\^{a} \vec{u}_{k}\^{s})\big\}}
                    {\sum_{k'=1}^{T\^{w}} \text{exp}\big\{\vec{v}_{\text{sd}}\^{a} \vec{u}_{k'}\^{s}\big\}} 
       \\ \vec{x}\^{d} &= \sum_{k=1}^{T\^{s}}\alpha_{k}\^{s}   \vec{h}_{k}\^{s},      
\end{split}
\end{equation}
with $\vec{u}_{k}\^{s} = \text{tanh}\big( \boldsymbol{W}_{\text{sd}}\^{a} \vec{h}_{k}\^{s} + \vec{b}_{\text{sd}}\^{a} \big)$
and $\vec{x}\^{d}$ representing the \textit{document embeddings}. 
Trainable parameters are $\boldsymbol{\theta}_{\text{sd}}\^{ATT}$ $=$ 
$\{\boldsymbol{W}_{\text{sd}}\^{a} \in \real^{D_{\text{sd}}\^{a} \times 2 \cdot D_{\text{d}}},$  
$\boldsymbol{b}_{\text{sd}}\^{a}\in \real^{D_{\text{sd}}\^{a} \times 1 },$ 
$\vec{v}_{\text{sd}}\^{a} \in \real^{1 \times  D_{\text{sd}}\^{a}}\}$.
%


\subsection{Nonlinear Metric Learning}
We now feed the document embbedings $\vec{x}\^{d}$ 
into a fully-connected multilayer perceptron (MLP),
\begin{align}
    \label{eq:MLP}
   \vec{y} = \text{tanh}\big( \boldsymbol{W}^{\^{f}}  \vec{x}\^{d} + \vec{b}^{\^{f}} \big),
\end{align}
to obtain the $D_{\text{f}}$-dimensional \textit{document features}~$\vec{y}$~\cite{Hu14}.
The parameter set associated with the MLP is denoted with  $\boldsymbol{\theta}\^{MLP}$ = $\{\boldsymbol{W}^{\^{f}} 
                        \in \real^{D_{\text{f}} \times 2 \cdot D_{\text{d}}},$ $ 
                        \boldsymbol{b}\^{f}\in \real^{D_{\text{f}} \times 1 }
\}$. 
Given a pair of document features, $\vec{y}_i$ for $i\in \{1,2\}$ via Eq.~(\ref{eq:MLP}), 
we can measure the similarity of both documents by determining the \emph{Euclidean} distance,
\begin{align}
\label{eq:dist}
    d\big(\vec{y}_1, \vec{y}_2\big) 
            = \norm{\vec{y}_1-\vec{y}_2}.
\end{align}


\begin{figure*}[t!]
\centering
\scalebox{0.85}{
\begin{psfrags}
\newcommand{\n}{0.9}
\newcommand{\m}{0.8}
    \psfrag{i}[c][c][\n]{$d($}
    \psfrag{j}[c][c][\n]{$,$}
    \psfrag{k}[c][c][\n]{$)<\tau_s$}
    \psfrag{k1}[c][c][\n]{$)$}
    \psfrag{v}[c][c][\n]{$d($}
    \psfrag{w}[c][c][\n]{$,$}
    \psfrag{q}[c][c][\n]{$)>\tau_d$}
    \psfrag{z1}[c][c][\m]{after}
    \psfrag{z2}[c][c][\m]{training}
    \psfrag{n}[c][c][\m]{document written by author $A$}
    \psfrag{m}[c][c][\m]{document written by author $B$}
    \psfrag{aa}[c][c][\m]{cross-topic}
    \psfrag{bb}[c][c][\m]{same topic}
    \psfrag{x}[c][c][\n]{$\tau_d$}
    \psfrag{y}[c][c][\n]{$\tau_s$}
    \psfrag{pr}[c][c][\n]{Text Preprocessing, Sentence Segmentation, Tokenization, Word/Character-to-vector Mapping}
    \psfrag{doc}[c][c][\n]{document $A$}
    \psfrag{c1}[c][c][\n]{1D CNN}
    \psfrag{c2}[c][c][\n]{+}
    \psfrag{c3}[c][c][\n]{max-pooling}
    \psfrag{co}[c][c][\m]{\textit{concatenation}}
    \psfrag{in}[c][c][\m]{\textit{interpolation}}
    \psfrag{sc}[c][c][\n]{scoring value $d(\vec{y}_1, \vec{y}_2)$}
    \psfrag{wsf}[c][c][\n]{$\overrightarrow{\text{LSTM}}_{\text{ws}}$}
    \psfrag{wsb}[c][c][\n]{$\overleftarrow{\text{LSTM}}_{\text{ws}}$}
    \psfrag{sdf}[c][c][\n]{$\overrightarrow{\text{LSTM}}_{\text{sd}}$}
    \psfrag{sdb}[c][c][\n]{$\overleftarrow{\text{LSTM}}_{\text{sd}}$}
    \psfrag{at0}[c][c][\n]{\,\,Max-over-time Pooling}
    \psfrag{at1}[c][c][\n]{$\text{Attention}_{\text{ws}}$}
    \psfrag{a21}[c][c][\n]{$\text{Attention}_{\text{sd}}$}
    \psfrag{d}[c][c][\n]{MLP}
    \psfrag{l}[c][c][\n]{$\ldots$}
    \psfrag{xd}[c][c][\n]{document embedding $ \vec{x}\^{d}$}
    \psfrag{er1}[c][c][\n][-90]{pseudo-metric learning}
    \psfrag{er2}[c][c][\n][-90]{}
    \psfrag{er3}[l][l][\n]{}
    \psfrag{er4}[c][c][\n][-90]{feature extraction}
    \psfrag{er5}[l][l][\n]{}
    \psfrag{er6}[l][l][\n]{}
    \psfrag{se}[c][c][\n]{sentence embeddings  $\vec{x}_{k}\^{s}$}
    \psfrag{we}[c][c][\n]{word embedding $\vec{x}_{j,k}\^{w}$}
    \psfrag{ce}[c][c][\n]{character embeddings $\vec{x}\^{c}_{i, j, k}$}
    \psfrag{cr}[c][c][\n]{character representation $\boldsymbol{r}_{j,k}$}
    \psfrag{y1}[c][c][\n]{document features $\vec{y}_1$, $\vec{y}_2$}
    \psfrag{grl1}[c][c][\n]{gradient}
    \psfrag{grl2}[c][c][\n]{reversal}
    \psfrag{grl3}[c][c][\n]{layer}
    \psfrag{ekl}[c][c][\n]{\,\,Distance Measure}
    \psfrag{dis}[c][c][\n]{$d(\vec{y}_1, \vec{y}_2)$}
    \psfrag{yn}[c][c][\n]{\textbf{yes} or \textbf{no} (and \textbf{fairly certain vs. uncertain})}
    \psfrag{pc}[c][c][\n]{$\Pr(\text{class})$}
    \psfrag{thr}[c][c][\n]{$~ 
                                \underset{\text{same author}}{\overset{\text{different authors}}{\gtrless}} \tau$}
    \psfrag{nw}[c][c][\n]{word/character embeddings of $j$-th word in $k$-th sentence}
    \psfrag{co}[c][c][\m]{\textit{concatenation}}
    \psfrag{aw}[c][c][\n]{word attentions $\alpha_{j,k}\^{w}$}
    \psfrag{as}[c][c][\n]{sentence attentions $ \alpha_{k}\^{s}$}
    %
    \includegraphics[width=14.5cm]{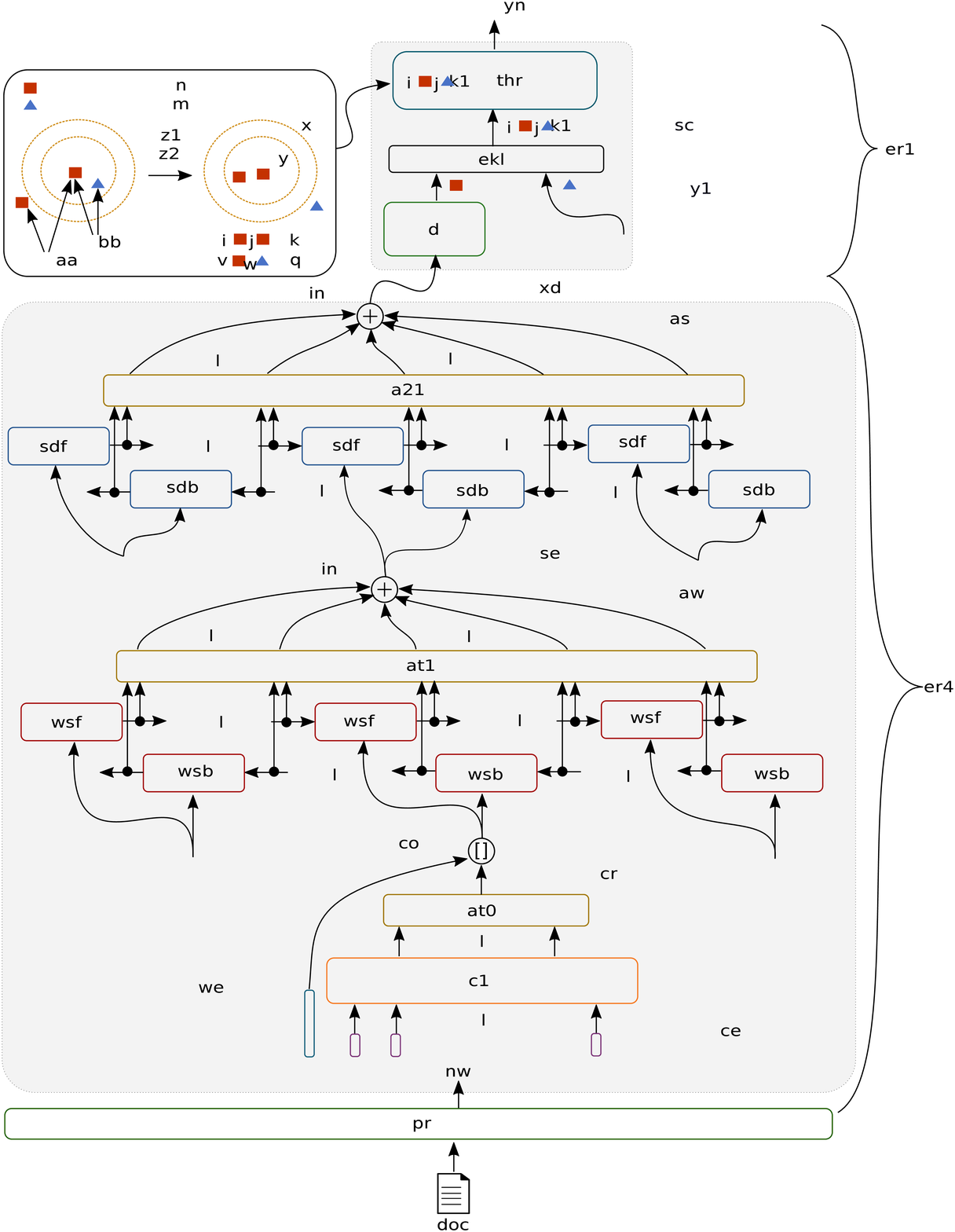}
\end{psfrags}
}
\caption{Architecture of the proposed \textsc{AdHominem} system. The feature extraction system for document $B$ (blue triangle) is not visible, since it is identical with the feature extraction system for document $A$ (red square).}
    \vspace*{-.5cm}
\label{fig:sn_att}
\end{figure*}

\subsection{The Loss Function}
\label{sec:loss}
The entire network is trained end-to-end, such that documents written by the same author should result in small values for Eq.~(\ref{eq:dist}), while the measure should return large values for documents of different authors.
A key problem in automatic text comparison is that it is generally much easier to compare the topical content of two documents than it is to compare their authorship. It is, thus, important that feature vectors in Eq.~(\ref{eq:MLP}) are made insensitive to topical variations between texts. A practical means to accomplish this insensitivity has been proposed in~\cite{Hu14, HRSN}. The approach is based on a {\em double threshold\/} concept illustrated in the top left corner of Figure~\ref{fig:sn_att}. Two distance thresholds $\tau_s$ and $\tau_d$ are defined with \mbox{$\tau_s<\tau_d$.} During training, all distances between document pairs (\authorA,\authorA) that are considered to belong to a same-author category (labeled with $a=1$) are to stay below the lower of the two thresholds, $\tau_s$, e.g.
\begin{align}
    \loss\^{s}_{\boldsymbol{\Theta}} = a \cdot \max \left\{ d\big(\vec{y}_1, \vec{y}_2\big) - \tau_s, 0 \right\}^2,
\end{align}
with all trainable parameters
$\boldsymbol{\Theta} =$ $\{\boldsymbol{\theta}\^{CNN},$ $\boldsymbol{\theta}_{\text{ws}}\^{fRNN},$ $\boldsymbol{\theta}_{\text{ws}}\^{bRNN},$ $\boldsymbol{\theta}_{\text{ws}}\^{ATT},$ $\boldsymbol{\theta}_{\text{sd}}\^{fRNN},$ 
$\boldsymbol{\theta}_{\text{sd}}\^{bRNN},$ $\boldsymbol{\theta}_{\text{sd}}\^{ATT},$ $\boldsymbol{\theta}\^{MLP}\}$.
Conversely, distances between document pairs (\authorA,\authorB) that belong to the different-author category (labeled with $a=0$) are to remain above the higher threshold $\tau_d$, e.g.
\begin{align}
    \loss\^{d}_{\boldsymbol{\Theta}} = (1-a) \cdot \max \left\{ \tau_d - d\big(\vec{y}_1, \vec{y}_2\big) , 0\right\}^2.
\end{align}
The final loss function is then given by 
\begin{align}
\label{Eq.combloss}
\loss_{\boldsymbol{\Theta}} = \loss\^{s}_{\boldsymbol{\Theta}} + \loss\^{d}_{\boldsymbol{\Theta}}.
\end{align}
Note that the loss function in Eq.~(\ref{Eq.combloss}) itself does not directly deal with the problem of topical variations.
Our motivation behind the double-threshold-mechanism is that algorithms generally tend to misclassify short texts annotated with \textit{same-author}/\textit{cross-topics} or \textit{different-authors}/\textit{same-topic}. 
 We therefore suggest to combine the loss function in Eq.~(\ref{Eq.combloss}) with a large, balanced dataset, where we have the {\em same\/} number of occurrences of \textit{same-topic} and \textit{cross-topic} cases to deal with. Document pairs that are easy to verify (\textit{same-author}/\textit{same-topic} and \textit{different-author}/\textit{cross-topic}) are ignored during training when their distances are under ($\le \tau_s$) or above ($\ge \tau_d$) the corresponding threshold. As a result, our system strongly focuses on more difficult document pairs. 

\section{\mbox{Large-scale Corpus for Text Comparison}}
\label{sec:dataacquisition}
We prepared a new large-scale corpus of short Amazon reviews as follows~\cite{McAuley15}:
\vspace*{-0.10in}
\begin{itemize}[leftmargin=5mm, noitemsep]
 \item Preprocessing steps are applied as discussed in Section~\ref{sec:impldet}. 
        All reviews with less then $80$ tokens and more than $1000$ tokens were removed. As a result, we obtained $9,052,606$ reviews written by $784,649$ authors, where, on average, each review consists of $282.10 \pm 198.14$ tokens.
        Additionally, the reviews for each author are grouped into $24$ different categories (by their meta-data)\footnote{Raw dataset available at \url{http://jmcauley.ucsd.edu/data/amazon}} to be able to analyze cross-topic/same topic instances.
 \item We removed all rare token types with less than $20$ overall occurrences to reduce the vocabulary size from $7,427,762$ to $317,712$ tokens. Analogously, we also removed all character types with less $100$ overall occurrences to reduce the size from $1,482$ to $222$ characters.
 \item In addition to \textit{zero-padding} tokens (for short sentences) and \textit{unknown} tokens, we also introduced tokens to deal with long sentences. If a sentence is shorter than a 
       predefined maximum sentence length then it ends with a regular \textit{sentence-ends} token. If a sentence is longer than the maximum sentence length then we stop with a \textit{line-break} token and shift the remaining part of the sentence into the next line.
\item For cross-validation, the reviews are randomly split into disjoint groups w.r.t. the authorship. As a result, development and test sets only contain unseen reviews written by unseen authors.
\item We sample review pairs w.r.t. to the authorship and category. Each review pair is labeled by a tuple 
        $l= (a,c)$ with $a\in\{0, 1\}$ and $c\in\{0, 1\}$. The value of $a$ indicates if the reviews were written by the same author ($a=1$) 
        or by different authors ($a=0$). The value of $c$ indicates if the reviews treat the same topic ($c=1$) or treat different topics ($c=0$).
\item The review pairs are recombined after each epoch 
        in order to increase the heterogeneity of the training set.
\item Each author contributes with only a minimum number of documents w.r.t. each tuple category $l= (a,c)$ which are represented equally to yield a balanced dataset.
\end{itemize}
\vspace*{-0.10in}
Altogether we obtain around $335,000$ training pairs for each epoch and $42,000$ instances for the development/test sets each, with an equal number of instances for all labels $l=(a,c)$.

\section{Evaluation}
\label{sec:Largescaleperformance}
\begin{figure}[t!]
        \centerline{\includegraphics[width=5.6cm]{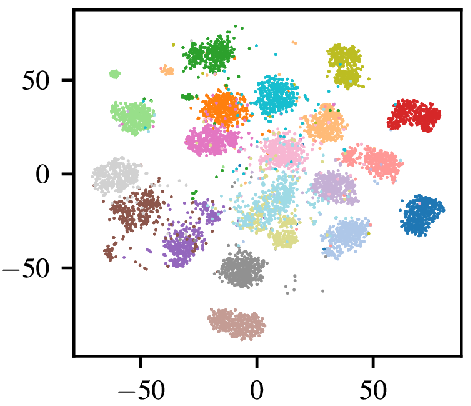}}
    \caption{A t-SNE-plot of neural features after~\cite{tsne}. We randomly selected $500$ reviews written by $20$ \textit{unseen} authors. Each of the color belongs to another author. The representation illustrates the discriminative power of \textsc{AdHominem} over unseen data.}
\label{fig:tsne_plot}
\vspace*{-0.5cm}
\end{figure}

In this section, we present experimental results for the proposed \textsc{AdHominem} system in comparison to a few other published approaches. A detailed documentation of the training procedure, including hyper-parameter settings, program code, and the data are accessible
\footnote{\url{https://github.com/rub-ksv/AdHominem}} 
for other researchers in the field.

\subsection{Implementation Details}
\label{sec:impldet}
\textsc{AdHominem} was implemented in Python. 
We utilized the library \verb|textacy|\footnote{\url{https://chartbeat-labs.github.io/textacy}} for data preprocessing.
All URLs, email addresses, and phone numbers were replaced with respective universal tokens as they themselves are typically not part of an author's writing style
but only the {\em use of them\/} is part of an author's writing style. The package \verb|spaCy|\footnote{\url{https://spacy.io/}} was used for sentence boundary detection and tokenization. 
The training of the neural networks was accomplished with \verb|Tensorflow|\footnote{\url{https://www.tensorflow.org/}}. 
Pretrained word embeddings were taken from \verb|fastText|~\cite{mikolov2018advances}. 

\begin{table}[t!]
\caption{Average verification error rates and standard deviations for the test set over a $5$-fold cross-validation.}
\centering{ 
\setstretch{1.3}
\small
\setlength\tabcolsep{3pt}
\noindent
\begin{tabular}{p{.9cm} p{1.3cm} p{1.3cm} p{1.3cm} p{1.3cm} p{1.35cm}} 
            \hline
           \multirow{ 2}{*}{\scalebox{.8}{labels}}      
            &\multicolumn{5}{c}{\scalebox{.8}{error rate in \%}}
                                \\
                        &\scalebox{.8}{\textsc{Imposters}} 
                        &\scalebox{.8}{\textsc{AvEeer}} 
                        &\scalebox{.8}{\textsc{Glad}} 
                        &\scalebox{.8}{\textsc{Hrsn}}
                        &\hspace*{-0.3em} \scalebox{.8}{\textsc{AdHominem}}
                        \\\arrayrulecolor{gray} \hline
    \scalebox{.8}{$\forall l$=$(a,c)$}   
        &\scalebox{.8}{$33.33 \pm 0.16$}    &\scalebox{.8}{$30.13\pm 0.13$}  
        &\scalebox{.8}{$27.13 \pm 0.14$}    &\scalebox{.8}{$15.40 \pm 0.19$}  
        &\scalebox{.8}{$14.70 \pm 0.16$}  
        \\  \hline
    \scalebox{.8}{$a$=$1$, $c$=$1$}   
        &\scalebox{.8}{$25.76 \pm 0.59$}    &\scalebox{.8}{$25.63 \pm 0.25$} 
        &\scalebox{.8}{$24.27 \pm 0.23$}    &\scalebox{.8}{$11.74 \pm 0.12$}      
        &\scalebox{.8}{$11.13 \pm 0.52$} 
        \\ \hline
    \scalebox{.8}{$a$=$1$, $c$=$0$}   
        &\scalebox{.8}{$40.27 \pm 1.13$}    &\scalebox{.8}{$35.88 \pm 0.53$} 
        &\scalebox{.8}{$35.97 \pm 0.57$}    &\scalebox{.8}{$17.81 \pm 0.33$}
        &\scalebox{.8}{$16.74 \pm 0.75$}    
        \\ \hline
    \scalebox{.8}{$a$=$0$, $c$=$1$}   
        &\scalebox{.8}{$42.54 \pm 0.94$}    &\scalebox{.8}{$36.37 \pm 0.54$} 
        &\scalebox{.8}{$30.28 \pm 0.54$}    &\scalebox{.8}{$22.20 \pm 0.65$}  
        &\scalebox{.8}{$21.24 \pm 1.03$}    
        \\ \hline
    \scalebox{.8}{$a$=$0$, $c$=$0$}   
        &\scalebox{.8}{$24.87 \pm 1.21$}    &\scalebox{.8}{$22.64 \pm 0.37$} 
        &\scalebox{.8}{$18.02 \pm 0.41$}    &~~\scalebox{.8}{$9.85 \pm 0.45$}
        &~~\scalebox{.8}{$9.68 \pm 0.79$}
        \\ \arrayrulecolor{black} \hline
  \end{tabular}}
\label{tab:acc1}
\vspace*{-.2cm}
\end{table}
\begin{table}[t!]
\caption{Average verification error rates and standard deviations for the test set over a $5$-fold cross-validation.
Only results for which the systems reported a {\bf high reliability\/} are considered.}
\setstretch{1.3}
\small
\setlength\tabcolsep{3pt}
\noindent
\begin{tabular}{p{1.1cm} p{1.55cm} p{1.75cm} p{1.55cm} p{1.75cm}} 
            \hline
            \multirow{ 3}{*}{\scalebox{.9}{labels}}    
            & \multicolumn{4}{c}{\scalebox{.9}{$d(\boldsymbol{y}_1,\boldsymbol{y}_2) \!\le\! \tau_s$ 
                        and $d(\boldsymbol{y}_1,\boldsymbol{y}_2) \!\ge\! \tau_d$}}
                                \\
            & \multicolumn{2}{c}{\scalebox{.9}{\textsc{Hrsn}}}
                   & \multicolumn{2}{c}{\scalebox{.9}{\textsc{AdHominem}}}
                    \\
                        &\scalebox{.85}{error rate (\%)} 
                        &\scalebox{.85}{\# instances (\%)} 
                        &\scalebox{.85}{error rate (\%)} 
                        &\scalebox{.85}{\# instances (\%)} 
                        \\\arrayrulecolor{gray} \hline
    \scalebox{.9}{$\forall l$=$(a,c)$}   
        & \scalebox{.9}{$0.93 \pm 0.04$}   &\scalebox{.9}{$19.91 \pm 0.49$} 
        &\scalebox{.9}{$0.84 \pm 0.07$} &\scalebox{.9}{$21.30 \pm 0.41$}       \\ \hline
    \scalebox{.9}{$a$=$1$, $c$=$1$}
        & \scalebox{.9}{$1.15 \pm 0.33$}   &\scalebox{.9}{$13.08 \pm 0.38$} 
        &\scalebox{.9}{$1.05 \pm 0.17$} &\scalebox{.9}{$14.66 \pm 0.90$}       \\ \hline
    \scalebox{.9}{$a$=$1$, $c$=$0$}   
        & \scalebox{.9}{$3.12 \pm 0.43$}   &\scalebox{.9}{$12.07 \pm 0.61$} 
        &\scalebox{.9}{$2.74 \pm 0.49$} &\scalebox{.9}{$13.35 \pm 0.55$}       \\ \hline
    \scalebox{.9}{$a$=$0$, $c$=$1$}   
        & \scalebox{.9}{$0.99 \pm 0.20$}   &\scalebox{.9}{$19.60 \pm 0.60$} 
        &\scalebox{.9}{$0.81 \pm 0.20$} &\scalebox{.9}{$20.93 \pm 1.18$}       \\ \hline
    \scalebox{.9}{$a$=$0$, $c$=$0$}   
        & \scalebox{.9}{$0.07 \pm 0.02$}   &\scalebox{.9}{$34.89 \pm 1.06$} 
        &\scalebox{.9}{$0.09 \pm 0.03$} &\scalebox{.9}{$36.25 \pm 1.38$}        \\ \arrayrulecolor{black} \hline
  \end{tabular}
\label{tab:acc2}
\vspace*{-.5cm}
\end{table}

\subsection{Baseline Methods}
We chose four published authorship verification methods as comparison references for our proposed approach.
\textsc{AVeer}~\cite{Halvani16}, \textsc{Glad}~\cite{DBLP:conf/clef/HurlimannWBSN15} and~\textsc{Imposters}~\cite{doi:10.1002/asi.22954} are based on a traditional stylometric feature extraction. These three algorithms have been ranked first, second and third in a performance evaluation on a small-sized corpus of larger Amazon reviews conducted by~\cite{DBLP:journals/corr/abs-1901-00399}. 
In addition, we also considered our predecessor \textsc{Hrsn}~\cite{HRSN}.

\subsection{Results}
\label{sec:res}
\begin{figure*}[t]
    \begin{minipage}[b]{.24\linewidth}
        \centering
        \centerline{\includegraphics[width=4.1cm]{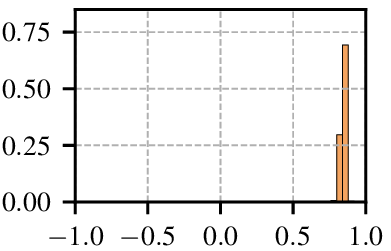}}
        \centerline{\small (a) mean/std: $0.846 \pm 0.015$}
    \end{minipage}
    \hfill
    \begin{minipage}[b]{.24\linewidth}
        \centering
        \centerline{\includegraphics[width=4.1cm]{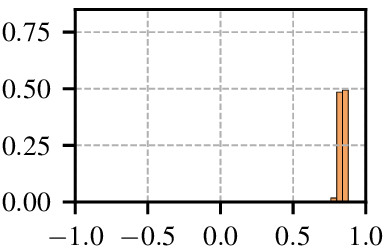}}
        \centerline{\small (b) mean/std: $0.838 \pm 0.017$}
    \end{minipage}
    \hfill
    \begin{minipage}[b]{.24\linewidth}
        \centering
        \centerline{\includegraphics[width=4.1cm]{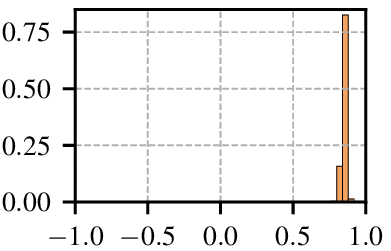}}
        \centerline{\small (c) mean/std: $0.853 \pm 0.014$}
    \end{minipage}
    \hfill
    \begin{minipage}[b]{.24\linewidth}
        \centering
        \centerline{\includegraphics[width=4.1cm]{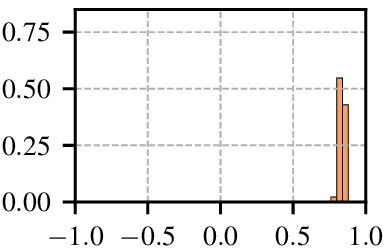}}
        \centerline{\small (d) mean/std: $0.836 \pm 0.016$}
    \end{minipage}
    \caption{Histograms of Kendall-$\tau$ correlation coefficients between weighted attentions of the reference run and all other cross-validations. 
         This correlation is statistically significant ($p\le 0.01$) for all instances.
         Values near $0$ indicate no correspondence and values close to $1$ imply perfect agreement.}
\label{fig:kendall_tau}
\vspace*{-.5cm}
\end{figure*}
Table~\ref{tab:acc1} summarizes the average verification error rates for the dataset described in Section~\ref{sec:dataacquisition} with a $5$-fold cross-validation. It is readily seen from the first row of Table~\ref{tab:acc1} that our two methods based on a Siamese topology (\textsc{Hrsn} and \textsc{AdHominem}) significantly outperformed the other three systems which are based on traditional stylometric features.
Comparing the results of the baseline methods \textsc{Imposters}, \textsc{Aveer} and \textsc{Glad}, we obtained error rates between $27\%-34\%$, while both, \textsc{Hrsn} and \textsc{AdHominem}, were able to cut the error rate in half to around $14\%-16\%$. Comparing \textsc{Hrsn} and \textsc{AdHominem}, we were able to slightly increase the average accuracy with our new attention-based Siamese topology.
Rows $2$-$5$ additionally show the performance w.r.t. the different label categories defined in Section~\ref{sec:dataacquisition}.
As it can be seen, all systems perform well for {\em same-author\/}/{\em same-category\/} instances as well as for {\em different-author\/}/{\em cross-topic\/} instances.
As one would expect, the error rates for all methods dramatically increase for {\em same-author\/}/{\em cross-topic\/} and {\em different-author\/}/{\em same-topic\/} cases. The results presented in Table~\ref{tab:acc1} show very clearly that the dataset discussed in Section~\ref{sec:dataacquisition} is quite challenging for all methods. The proposed Siamese network topologies, however, displayed a significantly higher discriminative power than the stylometric-feature-based systems. This is also evident in Fig.~\ref{fig:tsne_plot}, where, in a first step, we randomly selected $500$ documents written by $20$ different \textit{unseen} authors; and, in a second step, we computed the neural features $\vec{y}$ using the trained feature extraction module of \textsc{AdHominem} for each of the documents. In a third step, we applied a t-SNE to reduce the dimension to produce a visual representation. Each color in the plot belongs to a different author. As it can be seen, the features $\vec{y}$ produced by \textsc{AdHominem} are well suited to discriminate between different {\em unseen\/} authors.

In addition to overall error rate counts, we are also reporting the degree to which the Siamese network topologies are able to decide with a \textit{high level of reliability.} The \textit{double threshold} concept introduced in Section~\ref{sec:loss} can be leveraged to this end. If the feature vector tuples $(\vec{y}_1, \vec{y}_2)=(\text{\authorA},\text{\authorA})$ or $(\vec{y}_1, \vec{y}_2)=(\text{\authorA},\text{\authorB})$ of two texts have a distance $d(\vec{y}_1,\vec{y}_2)$ below $\tau_s$ or above $\tau_d$ then we may assume that the system attaches a high reliability to its decision. If the distance is between $\tau_s$ and $\tau_d$, however, then we may still arrive at a decision by comparing the distance to $\frac{\tau_s+\tau_d}{2}$, but the decision would carry much less ``confidence''. Results for text comparisons for which a {\em high level of reliability\/} is reported are presented in Table~\ref{tab:acc2}. 
We are separately reporting on the performance of \textsc{AdHominem} and \textsc{Hrsn} regarding cross-topic and same topic cases.
In Table~\ref{tab:acc2}, we observe that the error rates generally decrease dramatically when we only consider cases where the scores exceed our predefined thresholds. With both Siamese topologies we may reduce the error rate below $1\%$. It should be noted, however, that this holds for only a limited number of cases in our test set. The respective number of instances for which a {\em high reliability\/} is detected is reported in Table~\ref{tab:acc2} as well. The overall error rate, as reported in the first row of Table~\ref{tab:acc2}, is similar between \textsc{AdHominem} and \textsc{Hrsn}, but \textsc{AdHominem} improves slightly in the number of instances in which this is the case.

\subsection{Correlation Analysis}
In Section~\ref{sec:linguisticanalysis} we are providing a linguistic analysis of visualized attentions based on a single reference run from our 5-fold cross-validation. If the resulting attention weights bear merit as indicators of ``linguistic importance'' then we would expect that the weights obtained from different runs should be highly correlated. In order to study this correlation we formed an overall attention weight $\alpha_{j,k}$ as the product between the word-based attentions from Eq.~(\ref{eq:att_w}) and the associated sentence-based attentions from Eq.~(\ref{eq:att_s}), i.e. 
\begin{align}
        \label{eq:w_att}
        \alpha_{j,k} = \alpha\^{w}_{j,k}  \cdot \alpha\^{s}_{k}.
\end{align} 
We determined Kendall-$\tau$ coefficients of the $\alpha_{j,k}$ between the reference and all other cross-validation runs~\cite{AinE}. The results are shown in Fig.~\ref{fig:kendall_tau}. The weighted attentions $\alpha_{j,k}$ in Eq.~(\ref{eq:w_att})
from different cross-validation runs exhibit a high degree of correlation with the reference run. This supports our claim that attention weights may be used as indicators of ``linguistic importance''.

\subsection{Linguistic analysis}
\label{sec:linguisticanalysis}
\begin{figure}[t!]
    \vspace*{-0.6cm}
    \centering
    \begin{minipage}[t]{.48\linewidth}
        \centering
        \centerline{\includegraphics[width=4.11cm]{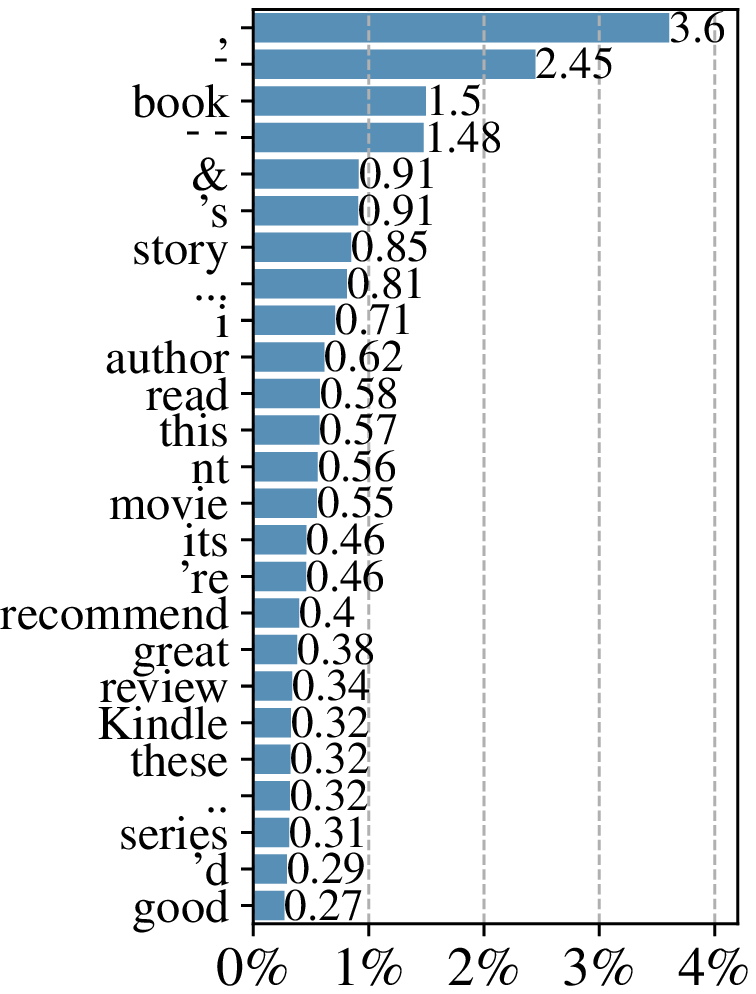}}
    \end{minipage}
    \begin{minipage}[t]{.48\linewidth}
        \centering
        \centerline{\includegraphics[width=4.35cm]{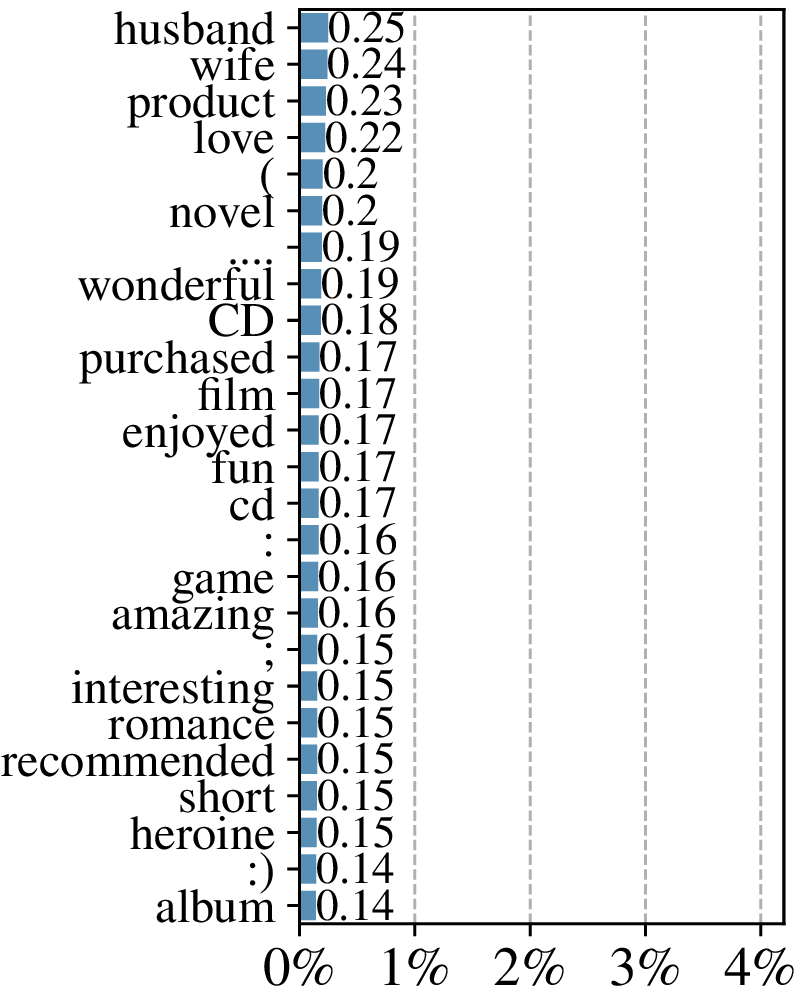}}
    \end{minipage}
    \caption{Frequencies of all tokens being listed in the top 5 ranking w.r.t the attention weights in a review.}
\label{fig:wf_plot}
\vspace*{-0.5cm}
\end{figure}
As already mentioned in the previous section, we were motivated by \cite{mullenbach-etal-2018-explainable} to check if any of the \textsc{AdHominem} attention weights would exhibit traits of linguistic categories. In order to support the analysis we utilize Eq.~(\ref{eq:w_att}) to color-code the texts as shown below, i.e.~a red background implies a high attention weight and a white background implies a low attention weight. 

Fig.~\ref{fig:wf_plot} provides a list of the collected top $5$ ranked tokens of each document that were tallied/counted across all documents. At the top of the list we only see \textit{known} tokens (for which we trained word embeddings), which is not surprising. Grammatical errors or mispellings are performed individually. However, our proposed system is able to handle \textit{unknown} tokens, e.g. the following found misspellings of the token \textit{definitely}: 
\vspace*{-0.06in}
\begin{displayquote}
\textit{diffenatly, definatley, definately, definitley, definitly, definetly, defintely, definatly, definalty, definantly, definetely, definitely(and, definently}
\end{displayquote}
\vspace*{-0.06in}
Cumulatively, they represent $0.13\%$ of the top 5 tokens. 

We examined 100 randomly chosen text comparisons and analyzed them in order to extract linguistic features. No linguistic category was excluded, but the analysis focused on those parts that had high attention weights. 
A limited number of examples of identified linguistic features, as highlighted by the system, is provided below.


%
\setlist{nolistsep}
\subsubsection*{Punctuation}
\begin{enumerate}[leftmargin=7mm,noitemsep]
 \item Some punctuation marks (e.g. hyphen, brackets, colon, comma) 
  are frequently marked: 
\vspace*{-0.06in}
\begin{testexample}[-]
\label{ex:1}
\footnotesize
{\setlength{\fboxsep}{0pt}\colorbox{white!0}{\parbox{0.95\textwidth}{
\colorbox{red!55.976856}{\strut Clarke} \colorbox{red!15.560889}{\strut is} \colorbox{red!19.585829}{\strut just} \colorbox{red!63.85777}{\strut satisfactory} \colorbox{red!100.0}{\strut -} \colorbox{red!4.623278}{\strut her} \colorbox{red!10.469007}{\strut Jane} \colorbox{red!13.072697}{\strut is} \colorbox{red!20.556332}{\strut too} \colorbox{red!22.50744}{\strut weak} \colorbox{red!11.8084345}{\strut and} \colorbox{red!54.407936}{\strut bland} \colorbox{red!11.589621}{\strut .} 
}}}
\end{testexample}
\item Accumulation of punctuation marks:
\vspace*{-0.06in}
\begin{testexample}[...]
\label{ex:3}
\footnotesize
{\setlength{\fboxsep}{0pt}\colorbox{white!0}{\parbox{0.95\textwidth}{
\colorbox{red!1.399315}{\strut I} \colorbox{red!15.523841}{\strut just} \colorbox{red!19.508127}{\strut read} \colorbox{red!2.8936138}{\strut the} \colorbox{red!16.59874}{\strut chapter} \colorbox{red!3.5990558}{\strut on} 
\colorbox{red!21.187895}{\strut Generosity} \colorbox{red!85.402695}{\strut ...} \colorbox{red!4.320949}{\strut and} \colorbox{red!3.1049767}{\strut it} \colorbox{red!8.056714}{\strut was} 
\colorbox{red!73.61441}{\strut PHENOMENAL} \colorbox{red!4.454424}{\strut !} 
 }}}
\end{testexample}
\item Special characters:
\vspace*{-0.06in}
\begin{testexample}[\&]
\footnotesize
{\setlength{\fboxsep}{0pt}\colorbox{white!0}{\parbox{0.95\textwidth}{
%
$[\ldots]$ \colorbox{red!6.448374}{\strut if} \colorbox{red!2.9923716}{\strut you} \colorbox{red!8.89952}{\strut need} \colorbox{red!4.3697605}{\strut to} \colorbox{red!8.454414}{\strut tweak} \colorbox{red!84.88666}{\strut \&} \colorbox{red!5.278907}{\strut bend} \colorbox{red!11.833664}{\strut them}
 }}}
\end{testexample}

\item Missing white spaces:
\vspace*{-0.06in}
\begin{testexample}[book.5 vs. books. 5]
\label{ex:2}
\footnotesize
{\setlength{\fboxsep}{0pt}\colorbox{white!0}{\parbox{0.95\textwidth}{
\colorbox{red!7.0476084}{\strut I} \colorbox{red!33.381676}{\strut highly} \colorbox{red!41.12426}{\strut recommend} \colorbox{red!27.384281}{\strut this} \colorbox{red!100.0}{\strut book.5} \colorbox{red!60.268414}{\strut Stars} 
 }}}
\end{testexample}
\end{enumerate}

\subsubsection*{Characters}
\begin{enumerate}[leftmargin=7mm,resume, noitemsep]

 \item Substitutions of characters (typing errors):
\vspace*{-0.06in}
 \begin{testexample}[deffinatly vs. definitely]
\footnotesize
{\setlength{\fboxsep}{0pt}\colorbox{white!0}{\parbox{0.95\textwidth}{
\colorbox{red!33.670025}{\strut Ice} \colorbox{red!28.375677}{\strut Cube} \colorbox{red!100.0}{\strut deffinatly} \colorbox{red!20.714401}{\strut has} \colorbox{red!11.590002}{\strut a} \colorbox{red!35.416412}{\strut style} \colorbox{red!17.097187}{\strut that} \colorbox{red!27.734137}{\strut is} \colorbox{red!42.74478}{\strut unimatatable} \colorbox{red!11.17133}{\strut .} 
 }}}
\end{testexample}
\item Missing characters:
\vspace*{-0.06in}
\begin{testexample}[clasics vs. classics]
\footnotesize
{\setlength{\fboxsep}{0pt}\colorbox{white!0}{\parbox{0.95\textwidth}{
%
$[\ldots]$ \colorbox{red!3.7045393}{\strut he} \colorbox{red!12.690079}{\strut went} \colorbox{red!4.874284}{\strut on} \colorbox{red!3.1936793}{\strut to} \colorbox{red!7.841366}{\strut put} \colorbox{red!6.1888175}{\strut out} 
\colorbox{red!5.9597535}{\strut some} \colorbox{red!16.632345}{\strut hip} \colorbox{red!25.333738}{\strut hop} \colorbox{red!91.45489}{\strut clasics} \colorbox{red!7.8705564}{\strut .} 
 }}}
\end{testexample}
 
\item Surplus (redundant) characters:
\vspace*{-0.06in}
\begin{testexample}[amazone vs. Amazon/amazon]
\footnotesize
{\setlength{\fboxsep}{0pt}\colorbox{white!0}{\parbox{0.95\textwidth}{
\colorbox{red!21.06968}{\strut i}
\colorbox{red!44.723793}{\strut searched} \colorbox{red!6.2457485}{\strut a} \colorbox{red!17.075426}{\strut lot} \colorbox{red!9.353463}{\strut of} \colorbox{red!13.974696}{\strut this} \colorbox{red!26.341042}{\strut kind} \colorbox{red!11.903786}{\strut of} \colorbox{red!22.515274}{\strut high} \colorbox{red!61.227978}{\strut tech} \colorbox{red!16.737242}{\strut light} 
\colorbox{red!11.434693}{\strut around} \colorbox{red!80.956788}{\strut amazone} 
 }}}
\end{testexample}

\end{enumerate}

\subsubsection*{Capitalization} 
\begin{enumerate}[leftmargin=7mm,resume, noitemsep]
\item Lower instead of upper case:
\vspace*{-0.06in}
\begin{testexample}[i vs. I]
\footnotesize
{\setlength{\fboxsep}{0pt}\colorbox{white!0}{\parbox{0.95\textwidth}{
$[\ldots]$
\colorbox{red!55.75782}{\strut i} \colorbox{red!63.924026}{\strut 'm} \colorbox{red!22.693174}{\strut glad} \colorbox{red!78.86678}{\strut i} \colorbox{red!47.49505}{\strut picked} \colorbox{red!37.40877}{\strut this} \colorbox{red!14.565041}{\strut .} 
 }}}
\end{testexample}

\item Upper instead of lower case:
\vspace*{-0.06in}
\begin{testexample}[Stars vs. stars]
\footnotesize
{\setlength{\fboxsep}{0pt}\colorbox{white!0}{\parbox{0.95\textwidth}{
\colorbox{red!7.0476084}{\strut I} \colorbox{red!33.381676}{\strut highly} \colorbox{red!41.12426}{\strut recommend} \colorbox{red!27.384281}{\strut this} \colorbox{red!100.0}{\strut book.5} \colorbox{red!60.268414}{\strut Stars} 
 }}}
\end{testexample}

\item Continuous capitalization:
\vspace*{-0.06in}
\begin{testexample}[TOTALLY AWESOME]
\footnotesize
{\setlength{\fboxsep}{0pt}\colorbox{white!0}{\parbox{0.95\textwidth}{
\colorbox{red!78.47545}{\strut Uncaged} \colorbox{red!12.852542}{\strut is} \colorbox{red!48.432808}{\strut TOTALLY} \colorbox{red!43.894836}{\strut AWESOME} \colorbox{red!2.8109434}{\strut !} \colorbox{red!0.80252206}{\strut !} \colorbox{red!0.59935206}{\strut !} \colorbox{red!0.5255825}{\strut !} \colorbox{red!0.6737272}{\strut !} 
}}}
\end{testexample}

\end{enumerate}

\subsubsection*{Compound and separate spelling}
\begin{enumerate}[leftmargin=7mm,resume, noitemsep]

\item Faulty compound spelling:
\vspace*{-0.06in}
\begin{testexample}[ripoff vs. rip off/rip-off]
\footnotesize
{\setlength{\fboxsep}{0pt}\colorbox{white!0}{\parbox{0.95\textwidth}{
%
$[\ldots]$
 \colorbox{red!43.439777}{\strut universe} \colorbox{red!4.193709}{\strut of} \colorbox{red!100.0}{\strut camelot} \colorbox{red!6.325944}{\strut that} \colorbox{red!12.926203}{\strut is} \colorbox{red!7.5134478}{\strut a} 
\colorbox{red!87.915054}{\strut disney} 
\colorbox{red!63.49571}{\strut ripoff} 
$[\ldots]$
%
   }}}
\end{testexample}

\item Faulty separate spelling:
\vspace*{-0.06in}
\begin{testexample}[story teller vs. storyteller]
\footnotesize
{\setlength{\fboxsep}{0pt}\colorbox{white!0}{\parbox{0.95\textwidth}{
\colorbox{red!14.74288}{\strut JJ} \colorbox{red!23.384834}{\strut Knight} \colorbox{red!9.516212}{\strut is} \colorbox{red!12.904051}{\strut an} \colorbox{red!32.455017}{\strut awesome} \colorbox{red!32.343952}{\strut story} \colorbox{red!59.267628}{\strut teller}
$[\ldots]$
%
   }}}
\end{testexample}

\end{enumerate}

\subsubsection*{Acronyms and abbreviations}
\begin{enumerate}[leftmargin=7mm,resume, noitemsep]

\item Usage of acronyms:
\vspace*{-0.06in}
\begin{testexample}[OMG]
\footnotesize
{\setlength{\fboxsep}{0pt}\colorbox{white!0}{\parbox{0.95\textwidth}{
\colorbox{red!51.74368}{\strut OMG} \colorbox{red!6.000225}{\strut ,} \colorbox{red!10.658193}{\strut someone} \colorbox{red!16.369366}{\strut finally} \colorbox{red!20.50121}{\strut figured} \colorbox{red!6.9728246}{\strut it} \colorbox{red!20.41306}{\strut out} \colorbox{red!8.927758}{\strut !} 

   }}}
\end{testexample}

\item Abbreviations without punctuation marks:
\vspace*{-0.06in}
\begin{testexample}[Mr Rochester vs. Mr. Rochester]
\footnotesize
{\setlength{\fboxsep}{0pt}\colorbox{white!0}{\parbox{0.95\textwidth}{
%
$[\ldots]$
\colorbox{red!29.382002}{\strut performance} \colorbox{red!4.2722087}{\strut as} \colorbox{red!34.65375}{\strut Mr} \colorbox{red!11.006836}{\strut Rochester} \colorbox{red!13.176124}{\strut is} \colorbox{red!48.25917}{\strut superb}
$[\ldots]$
 %
   }}}
\end{testexample}

\item Unusual abbreviations:
\vspace*{-0.06in}
\begin{testexample}[def vs. definitely]
\footnotesize
{\setlength{\fboxsep}{0pt}\colorbox{white!0}{\parbox{0.95\textwidth}{
%
$[\ldots]$
\colorbox{red!5.7706895}{\strut and} \colorbox{red!3.5959172}{\strut that} \colorbox{red!10.163929}{\strut track} \colorbox{red!8.985845}{\strut is} \colorbox{red!54.96697}{\strut def} \colorbox{red!13.648321}{\strut tight} 
   }}}
\end{testexample}
\end{enumerate}

\subsubsection*{Diatopic variations and foreign languages}
\begin{enumerate}[leftmargin=7mm,resume, noitemsep]

\item British English vs. American English:
\vspace*{-0.06in}
\begin{testexample}[favourite (BE) vs. favorite (AE)]
\footnotesize
{\setlength{\fboxsep}{0pt}\colorbox{white!0}{\parbox{0.95\textwidth}{
$[\ldots]$
\colorbox{red!12.039204}{\strut this} \colorbox{red!19.857153}{\strut version} 
\colorbox{red!8.992147}{\strut is} \colorbox{red!35.2162}{\strut undoubtedly}
\colorbox{red!3.9770927}{\strut my} \colorbox{red!71.30594}{\strut favourite} \colorbox{red!9.021111}{\strut .} 
   }}}
\end{testexample}

\item Foreign words:
\vspace*{-0.06in}
\begin{testexample}[cloisonne (French: cloisonn\'e vs. Cloisonn\'e)]
\footnotesize
{\setlength{\fboxsep}{0pt}\colorbox{white!0}{\parbox{0.95\textwidth}{
 \colorbox{red!2.7965}{\strut I} \colorbox{red!20.5677}{\strut 've} \colorbox{red!17.7040}{\strut taken} \colorbox{red!4.0893}{\strut a} \colorbox{red!9.0771}{\strut few} \colorbox{red!19.7141}{\strut cloisonne} \colorbox{red!16.1531}{\strut classes}
$[\ldots]$
   }}}
\end{testexample}

\end{enumerate}

\subsubsection*{Stylistic features}
\begin{enumerate}[leftmargin=7mm,resume, noitemsep]

\item Unusual discourse particles/interjections:
\vspace*{-0.06in}
\begin{testexample}[hah]
\footnotesize
{\setlength{\fboxsep}{0pt}\colorbox{white!0}{\parbox{0.95\textwidth}{
 \colorbox{red!16.003956}{\strut What} \colorbox{red!60.79657}{\strut author} \colorbox{red!8.359511}{\strut or} \colorbox{red!54.883556}{\strut editor} \colorbox{red!15.14282}{\strut would} \colorbox{red!32.42262}{\strut let} \colorbox{red!6.6614375}{\strut that} \colorbox{red!18.621733}{\strut go} \colorbox{red!9.5020075}{\strut to} 
  \colorbox{red!27.561161}{\strut print?"Ah} 
  
  \colorbox{red!4.011089}{\strut -} \colorbox{red!43.822983}{\strut hum} \colorbox{red!5.807184}{\strut "} \colorbox{red!4.0209284}{\strut ,} \colorbox{red!4.266561}{\strut "} \colorbox{red!7.7246213}{\strut ah}
 \colorbox{red!1.9070969}{\strut -} \colorbox{red!18.099579}{\strut huh} \colorbox{red!5.878349}{\strut "} \colorbox{red!4.4059443}{\strut and} \colorbox{red!3.9786391}{\strut "} \colorbox{red!8.837623}{\strut a} \colorbox{red!9.181494}{\strut -} 
\colorbox{red!80.44152}{\strut hah} \colorbox{red!10.258011}{\strut "} 
$[\ldots]$
  %
   }}}
\end{testexample}

\item Colloquial expressions/slang:
\vspace*{-0.06in}
\begin{testexample}[thingamajig]
\footnotesize
{\setlength{\fboxsep}{0pt}\colorbox{white!0}{\parbox{0.95\textwidth}{
%
$[\ldots]$
\colorbox{red!10.222759}{\strut percentage} \colorbox{red!2.2743247}{\strut at} \colorbox{red!1.7229815}{\strut the} \colorbox{red!3.8894427}{\strut bottom} \colorbox{red!52.922947}{\strut thingamajig}
   }}}
\end{testexample}

\item Alternative spelling:
\vspace*{-0.06in}
\begin{testexample}[frikkin vs. freaking]
\footnotesize
{\setlength{\fboxsep}{0pt}\colorbox{white!0}{\parbox{0.95\textwidth}{
%
$[\ldots]$
\colorbox{red!9.658571}{\strut ,} \colorbox{red!40.585854}{\strut its} \colorbox{red!7.5981364}{\strut a} \colorbox{red!73.43885}{\strut frikkin} \colorbox{red!36.073517}{\strut tv} \colorbox{red!34.88372}{\strut show} \colorbox{red!12.085024}{\strut .} 
   }}}
\end{testexample}

\item Neologisms:
\vspace*{-0.06in}
\begin{testexample}[cartoonish]
\footnotesize
{\setlength{\fboxsep}{0pt}\colorbox{white!0}{\parbox{0.95\textwidth}{
\colorbox{red!17.187944}{\strut But} \colorbox{red!1.5504795}{\strut ,} \colorbox{red!35.155914}{\strut Lily} \colorbox{red!39.45268}{\strut feels} \colorbox{red!100.0}{\strut cartoonish} \colorbox{red!7.738858}{\strut to} \colorbox{red!20.725124}{\strut me} \colorbox{red!25.005222}{\strut ,} 
$[\ldots]$
%
    %
}}}
\end{testexample}

\end{enumerate}

\subsubsection*{Syntax}
\begin{enumerate}[leftmargin=7mm,resume, noitemsep]

\item Verb in first position (declarative sentence), missing noun or pronoun:
\vspace*{-0.06in}
\begin{testexample}[[I] Will]
\footnotesize
{\setlength{\fboxsep}{0pt}\colorbox{white!0}{\parbox{0.95\textwidth}{
\colorbox{red!34.82867}{\strut Will} \colorbox{red!48.628033}{\strut definitely} \colorbox{red!37.011192}{\strut read} \colorbox{red!16.236053}{\strut more} \colorbox{red!10.518501}{\strut of} \colorbox{red!18.550095}{\strut her} \colorbox{red!58.897827}{\strut books} \colorbox{red!12.008331}{\strut .} 
}}}
\end{testexample}
\end{enumerate} 

\subsubsection*{Proper nouns}
\begin{enumerate}[leftmargin=7mm,resume, noitemsep]

\item Proper nouns:
\vspace*{-0.06in}
\begin{testexample}[Mrs. Kuklinski]
\footnotesize
{\setlength{\fboxsep}{0pt}\colorbox{white!0}{\parbox{0.95\textwidth}{
\colorbox{red!21.028093}{\strut Winona} \colorbox{red!26.00779}{\strut Ryder} \colorbox{red!17.999203}{\strut also} \colorbox{red!12.579785}{\strut did} \colorbox{red!8.563449}{\strut a} \colorbox{red!24.443438}{\strut good} \colorbox{red!33.95415}{\strut job} \colorbox{red!8.98726}{\strut as} \colorbox{red!36.547775}{\strut Mrs} \colorbox{red!57.321815}{\strut Kuklinski} \colorbox{red!13.782061}{\strut .} 
}}}
\end{testexample}
  
\item Proper nouns vs. determinative compounds:
\vspace*{-0.06in}
\begin{testexample}[superman vs. Superman]
\footnotesize
{\setlength{\fboxsep}{0pt}\colorbox{white!0}{\parbox{0.95\textwidth}{
%
$[\ldots]$
\colorbox{red!6.3150177}{\strut gunshot} \colorbox{red!3.4036121}{\strut wounds} \colorbox{red!5.5691543}{\strut ca} \colorbox{red!5.7238092}{\strut n't} \colorbox{red!3.3500023}{\strut even} 
\colorbox{red!4.9327145}{\strut stop} \colorbox{red!3.0172846}{\strut our} \colorbox{red!35.233925}{\strut superman} \colorbox{red!4.2994905}{\strut .} 
}}}
\end{testexample}
\end{enumerate}

\subsubsection*{Additional features} 
\begin{enumerate}[leftmargin=7mm,resume, noitemsep]

\item Similar expressions:
\vspace*{-0.06in}
\begin{testexample}[straightforward vs.~straight forward]
\footnotesize
{\setlength{\fboxsep}{0pt}\colorbox{white!0}{\parbox{0.95\textwidth}{
 \colorbox{red!7.4696684}{\strut It} \colorbox{red!10.076061}{\strut was} \colorbox{red!35.781765}{\strut clear} \colorbox{red!9.844402}{\strut and} \colorbox{red!90.672714}{\strut straightforward} 
 $[\ldots]$
}}}
\end{testexample}

\item Combination of nouns and digits:
\vspace*{-0.06in}
\begin{testexample}[7-year]
\footnotesize
{\setlength{\fboxsep}{0pt}\colorbox{white!0}{\parbox{0.95\textwidth}{
%
$[\ldots]$
\colorbox{red!12.077479}{\strut with} \colorbox{red!6.8137913}{\strut my} 
\colorbox{red!77.251205}{\strut 7-year} \colorbox{red!44.033314}{\strut -} \colorbox{red!43.874023}{\strut old} \colorbox{red!8.546791}{\strut .} 
}}}
\end{testexample}

\item Systematic repetition of a mistake:
\vspace*{-0.06in}
\begin{testexample}[albumns vs. albums]
\footnotesize
{\setlength{\fboxsep}{0pt}\colorbox{white!0}{\parbox{0.95\textwidth}{
%
$[\ldots]$
\colorbox{red!7.289922}{\strut on} \colorbox{red!3.5507555}{\strut the} \colorbox{red!10.622755}{\strut live} \colorbox{red!83.751396}{\strut albumns} $[\ldots]$
\newline
%
 %
}}}
\end{testexample}
\end{enumerate}
\subsubsection*{Examples of combined linguistic features} 
\begin{enumerate}[leftmargin=7mm,resume, noitemsep]

\item Accumulated special characters:
\vspace*{-0.06in}
\begin{testexample}[A++++++]
\footnotesize
{\setlength{\fboxsep}{0pt}\colorbox{white!0}{\parbox{0.95\textwidth}{
\colorbox{red!19.583563}{\strut Mom} \colorbox{red!2.8008842}{\strut ,} \colorbox{red!3.8056295}{\strut you} \colorbox{red!10.403995}{\strut get} \colorbox{red!9.516816}{\strut an} \colorbox{red!50.244545}{\strut A++++++} \colorbox{red!6.200655}{\strut from} \colorbox{red!14.6091175}{\strut me} \colorbox{red!24.145536}{\strut ,} \colorbox{red!4.5028043}{\strut and} \colorbox{red!2.7880678}{\strut I} \colorbox{red!21.850286}{\strut am} 
}}}
\end{testexample}
\item Mispelled proper nouns:
\vspace*{-0.06in}
\begin{testexample}[Speilberg]
\footnotesize
{\setlength{\fboxsep}{0pt}\colorbox{white!0}{\parbox{0.95\textwidth}{
\colorbox{red!32.42364}{\strut Steven} \colorbox{red!94.998566}{\strut Speilberg} \colorbox{red!24.760992}{\strut 's} \colorbox{red!15.488265}{\strut first} \colorbox{red!30.0902}{\strut movie} \colorbox{red!6.6704383}{\strut as} \colorbox{red!23.507706}{\strut director} 
$[\ldots]$
%
}}}
\end{testexample}
\end{enumerate}

The analysis of linguistic features for the purpose of author verification is a complex research field, 
 since there are many levels of description, such as syntax, morphology, lexicology, semantics, pragmatics, spelling etc. Concepts from corpus linguistics and variational linguistics are also significant for a comprehensive analysis. 
 Our investigation, however, is limited by the algorithm’s estimation of the significance of terms and other parts of the texts. 
In applying an in-depth analysis of, e.g., syntactic features, whole sentences should be taken into account. When the algorithms flags 
an entire sentence, in turn, it is impossible to automatically determine which feature caused that decision. 
In addition to focusing on (topic-related) nouns, adjectives and verbs, which is also visible in Fig.~\ref{fig:wf_plot}, the algorithm favors some parts of the text over others; i.e., the last sentence of a review is frequently marked by the system. 

\subsection{Differential Example and Adversarial Example}
Lastly, we want to illustrate a conceptual feature of the proposed \textsc{AdHominem} system by returning to our example from Fig.~\ref{fig:example}. The \textsc{AdHominem} system predicts, correctly, that both reviews from Fig.~\ref{fig:example} were written by the same author. However, the decision is wrought with low reliability because $\tau_s=1<d(\boldsymbol{y}_1, \boldsymbol{y}_2)=1.66<\tau_d=3$. The resulting distance of $1.66$ is still below the threshold of $\frac{\tau_s+\tau_d}{2}=2$, which leads to the correct decision. The reason for the low reliability can be found in the lack of overlap in the token set observed in each of the reviews. The overlap is only $15.7\%$.




An interesting differential example is constructed if we substitute the words \textit{``cuz''} and \textit{``w/''} with \textit{``because''} and \textit{``with''} in the third sentence of the first review.
The updated attention-based heatmap, as produced by \textsc{AdHominem}, can be seen in Fig.~\ref{fig:difan}. As expected, our correction of the misspellings leads to a reduced attention for both words as well as for the entire sentence, which supports our claim that the chosen attentions are able to pick up on the relevance of individual linguistic features. The scoring slightly changes from $d(\boldsymbol{y}_1, \boldsymbol{y}_2)= 1.66$ to $d(\boldsymbol{\widetilde{y}_1}, \boldsymbol{y}_2)= 1.68$, so that the differential example would still be categorized correctly. 

Next, we compare reviews $4$ and $6$ in Fig.~\ref{fig:difan2}. Again, both reviews were written by the same author, with a similar overlap rate of $15.1\%$. Our system \textsc{AdHominem} predicted it correctly, with a score of $d(\boldsymbol{y}_1, \boldsymbol{y}_2)= 1.60$. It is observable that the author likes to use the 
character \textit{``\&''} instead of writing the word \textit{``and''}. We now create an adversarial example by replacing the $\&$ character in example 4 by the word \textit{``and''}.
The updated attention-based heatmap can be seen in example $5$ in Fig.~\ref{fig:difan2}. Again, \textsc{AdHominem} works as expected and the word \textit{``and''} is no longer marked as significant. As a result, however, the scoring increases to $d(\boldsymbol{\widetilde{y}_1}, \boldsymbol{y}_2)= 2.30$ which leads to a misclassification. 

Both examples exhibit the following properties: Although the review pairs have a very similar overlap rate, a simple manually performed error correction can lead to clear effects on the attention weights. On the one hand, if the reviews as in Fig.~\ref{fig:difan2} are too short (with $59$ and $63$ tokens, respectively) it is difficult to gather enough information for a reliable decision. As a result, \textsc{AdHominem} strongly relies on a single linguistic feature. This adversarial example shows that a very simple obfuscation strategy has the power to potentially impede correct verification,
an insight which can be leveraged in future work to better understand the decision-making process for automatic verification results and for hardening such frameworks systematically against obfuscation strategies.
On the other hand, considering Fig.~\ref{fig:example} and~\ref{fig:difan}, the verification of \textsc{AdHominem} is still robust. In both reviews, the number of tokens ($120$ and $130$) is sufficient to characterize the underlying writing styles. 

\begin{figure}[t]
    \vspace*{-.3cm}
\centering
\small
\begin{testexample2}[$\boldsymbol{\widetilde{y}}_1$]
        \scalebox{0.86}{
        \input{differentialanalysis/doc_L1.tex}
        }
\end{testexample2}
\label{fig:difana1}
\vspace*{-.1cm}
\caption{Differential example: scoring slightly changes from $d(\boldsymbol{y}_1, \boldsymbol{y}_2)= 1.66$ to $d(\boldsymbol{\widetilde{y}_1}, \boldsymbol{y}_2)= 1.68$.}
\label{fig:difan}
    \vspace*{-.5cm}
\end{figure}
\begin{figure}[t]
\vspace*{-.1cm}
\centering
\small
\begin{testexample2}[$\boldsymbol{y}_1$]
        \scalebox{0.86}{
        \input{differentialanalysis/doc_L2.tex}
        }
\end{testexample2}
\vspace*{-.2cm}
\begin{testexample2}[$\boldsymbol{\widetilde{y}}_1$]
        \scalebox{0.86}{
        \input{differentialanalysis/doc_L3.tex}
        }
\end{testexample2}
\vspace*{-.2cm}
\begin{testexample2}[$\boldsymbol{y}_2$]
        \scalebox{0.86}{
            \input{differentialanalysis/doc_R2.tex}
            }
\end{testexample2}
\label{fig:difana2}
\vspace*{-.1cm}
\caption{Adversarial example: scoring changes from $d(\boldsymbol{y}_1, \boldsymbol{y}_2)= 1.60$ to $d(\boldsymbol{\widetilde{y}_1}, \boldsymbol{y}_2)= 2.30$.}
\label{fig:difan2}
    \vspace*{-.5cm}
\end{figure}
%
%


\section{Conclusion}
\label{sec:conclusion}

We introduced a new algorithm for forensic text comparison called \textsc{AdHominem} which is characterized by an attention-based Siamese network topology that is able to learn linguistic features such as spelling errors, non-standard lexical forms, and expressions that differ in other ways from the norm. We compiled a large-scale dataset of short product reviews and made it publicly accessible for text comparison tasks. Besides the quantitative evaluation of the performance of the algorithm we presented a qualitative linguistic analysis based on the visualization of the internal attention weights.  

From a linguistic perspective, we grant that, despite our very encouraging results, further research may be needed. A limiting factor for any quantitative analysis in forensic linguistics is that, to our knowledge, no reliable mechanism exists to do the analysis by machine. This implies that all results that involve explainable features have to rely on a manual analysis by an expert and become, therefore, very difficult to do on large-scale datasets. Within the resources available to us we were therefore not able to conduct a comprehensive quantitative analysis of explainability. We, nevertheless, feel that our results were very encouraging and therefore worth sharing with researchers in the field.


From a technical perspective, our experiments highlight the distinct advantages of the proposed Siamese network topology over traditional methods.
\textsc{AdHominem} fuses a {\em self-configuring\/} feature extraction into a verification module to form a single framework. It is neither necessary, to manually define meaningful stylometric features, nor to acquire annotated data for preparatory steps like part-of-speech tagging.
Hence, the proposed framework allows for an efficient, fully automated use of big datasets in forensic linguistics, while still retaining a large degree of interpretability, which is of special significance in this field of application.
Our system demonstrates unequivocally that {\em big data\/} has become important and relevant in the field of forensic linguistic as well.





\bibliographystyle{IEEEtran}
\bibliography{adhominem}

\end{document}